\pdfoutput=1

\documentclass[final,5p,times,authoryear]{elsarticle}

\usepackage{amssymb}
\usepackage{amsmath}
\usepackage{url}
\usepackage{color}
\usepackage{hyperref}
\usepackage{booktabs}
\usepackage{multirow}

\journal{ISPRS Journal of Photogrammetry and Remote Sensing}

\begin{document}

\begin{frontmatter}



\title{iEBAKER: Improved Remote Sensing Image-Text Retrieval Framework via Eliminate Before Align and Keyword Explicit Reasoning}

\author[1]{Yan Zhang}
\ead{yzhang1995@tju.edu.cn}
\author[1]{Zhong Ji\corref{cor1}}
\ead{jizhong@tju.edu.cn}
\author[1]{Changxu Meng}
\ead{mengchangxu@tju.edu.cn}
\author[1]{Yanwei Pang}
\ead{pyw@tju.edu.cn}
\author[2]{Jungong Han}
\ead{jghan@tsinghua.edu.cn}

\address[1]{{School of Electrical and Information Engineering, Tianjin Key 
            Laboratory of Brain-inspired Intelligence Technology, Tianjin University},
            {No.92 Weijin Road}, 
            {Tianjin},
            {300072}, 
            {China}}

\address[2]{{Department of Automation, Tsinghua University},
            {No.30 Shuangqing Road}, 
            {Beijing},
            {100084}, 
            {China}}
            
\cortext[cor1]{Corresponding author}

\begin{abstract}
Recent studies focus on the Remote Sensing Image-Text Retrieval (RSITR), which aims at searching for the corresponding targets based on the given query. Among these efforts, the application of Foundation Models (FMs), such as CLIP, to the domain of remote sensing has yielded encouraging outcomes. However, existing FM based methodologies neglect the negative impact of weakly correlated sample pairs and fail to account for the key distinctions among remote sensing texts, leading to biased and superficial exploration of sample pairs. To address these challenges, we propose an approach named iEBAKER (an Improved Eliminate Before Align strategy with Keyword Explicit Reasoning framework) for RSITR. Specifically, we propose an innovative Eliminate Before Align (EBA) strategy to filter out the weakly correlated sample pairs, thereby mitigating their deviations from optimal embedding space during alignment.Further, two specific schemes are introduced from the  perspective of whether local similarity and global similarity affect each other. On this basis, we introduce an alternative Sort After Reversed Retrieval (SAR) strategy, aims at optimizing the similarity matrix via reverse retrieval. Additionally, we incorporate a Keyword Explicit Reasoning (KER) module to facilitate the beneficial impact of subtle key concept distinctions. Without bells and whistles, our approach enables a direct transition from FM to RSITR task, eliminating the need for additional pretraining on remote sensing data. Extensive experiments conducted on three popular benchmark datasets demonstrate that our proposed iEBAKER method surpasses the state-of-the-art models while requiring less training data. Our source code will be released at \textcolor{blue}{https://github.com/zhangy0822/iEBAKER}.
\end{abstract}




\end{frontmatter}



\section{Introduction}\label{sec1}
With the rapid progression of aerospace technology, remote sensing imagery has become increasingly accessible and is now extensively utilized in various fields, including disaster monitoring \citep{WANG2025139,li2020review}, navigation \citep{LI202425}, and agricultural production \citep{weiss2020remote}. Among these diverse applications, Remote Sensing Image-Text Retrieval (RSITR) emerges as a pivotal technique within the remote sensing vision-language domain \citep{zhao2025luojiahog}, with the goal of retrieving semantically similar images based on text queries, and conversely, identifying relevant text descriptions from image inputs.

\begin{figure}[t]
  \centering
  \includegraphics[width=0.90\columnwidth]{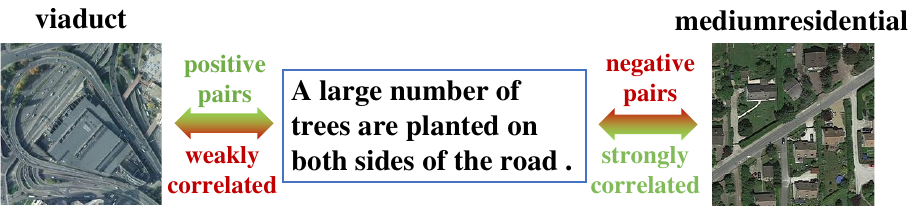}      
  \caption{Illustration of weakly correlated pairs. The left ``viaduct'' image is captioned as ``A large number of trees are planted on both sides of the road'', whereas they are weakly correlated. In contrast, the right ``a medium residential'' image is strongly correlated with above caption but considered as a negative image-text pair.}
  \label{example}
\end{figure}

The existing RSITR methods are mainly divided into traditional \citep{yuan2022mcrn,yuan2022remote,zhang2023hypersphere,ji2023knowledge} and 
Foundation Model (FM) based approaches \citep{liu2023remoteclip, zhang2023rs5m, zhong2024urbancross, yang2024accurate, ji2024eliminate}. 
Both approaches require meticulously curated datasets, and finely and accurately annotated RSITR data will contribute to improving the performance of the model \citep{zhang2023rs5m}. 
Nevertheless, despite rigorous annotation efforts, datasets may still contain irrelevant or weakly correlated image-text pairs \citep{tang2023interacting}. 
These weakly aligned pairs hinder the accurate alignment of semantically meaningful instances. For instance, as illustrated in Fig. \ref{example}, an image of a ``viaduct'' might be labeled as ``a large number of trees are planted on both sides of the road”, which is more pertinent to ``mediumresidential'' context,
offering little value to the model. 
Consequently, there is a critical need to explore mechanisms that allow the model to autonomously mitigate the adverse effects of such noise prior to engaging in fine-grained alignment.

\begin{figure*}[!t]
  \centering
  \includegraphics[width=0.95\textwidth]{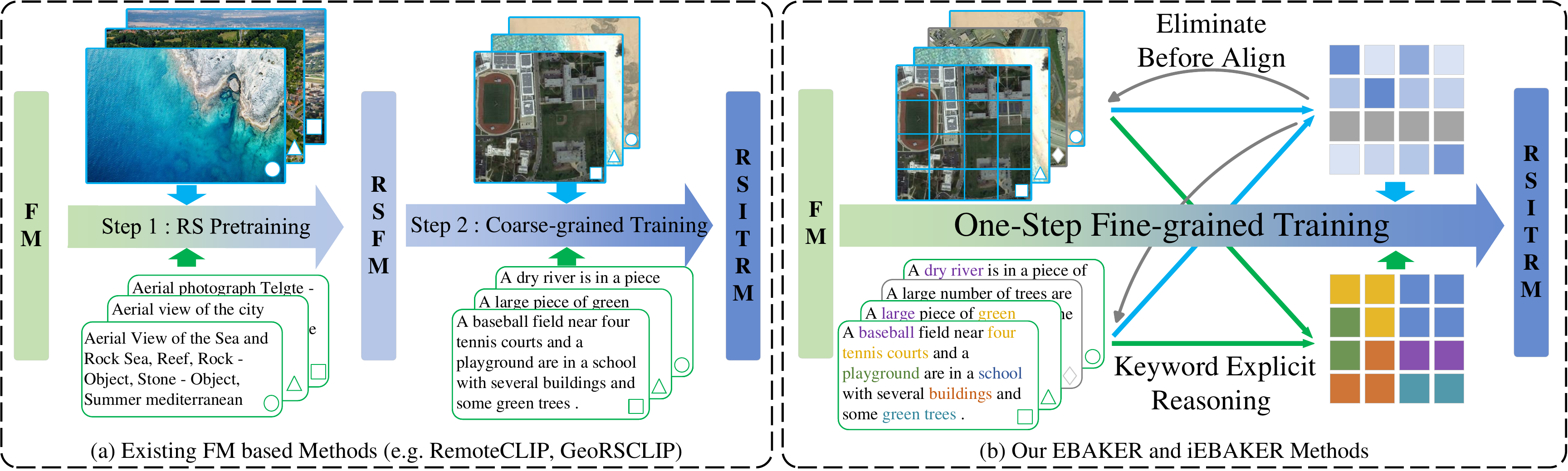}      
  \caption{Comparison between our EBAKER \citep{ji2024eliminate} and iEBAKER with the existing FM based methods on RSITR task. 
  (a) A significant volume of model-annotated remote sensing image-text pairs are employed to adapt the Foundation Model (FM) into a Remote Sensing Foundation Model (RSFM). Subsequently, the RSFM undergoes a further transformation into a Remote Sensing Image-Text Retrieval Model (RSITRM) through additional coarse-grained contrastive learning on the RSITR dataset. (b) In contrast, our approach achieves a direct one-step transition from FM to RSITRM by integrating the Eliminate Before Align (EBA) strategy and the Keyword Explicit Reasoning (KER) module, 
  streamlining the process and enhancing retrieval accuracy.}
  \label{motivation}
\end{figure*}

Moreover, while existing FMs have advanced RSITR by leveraging larger data volumes \citep{liu2023remoteclip,zhang2023rs5m}, they fail to address the core challenge of the task.The essence of RSITR lies not simply in expanding the quantity of positive and negative sample pairs for contrastive learning, but in thoroughly investigating the critical distinctions between these pairs. Both traditional methods and FM based approaches in RSITR tend to rely on global features generated by vision and text encoders, neglecting the key differences guided by the fine-grained features within remote sensing images.

These two challenges significantly limit the potential of FMs, despite their impressive advancements in various traditional downstream tasks, such as those demonstrated by CLIP \citep{radford2021learning} and BLIP \citep{li2022blip}.
Adapting FMs to Remote Sensing Image-Text Retrieval Model (RSITRM) still demands a considerable volume of remote sensing data, as depicted in Fig. \ref{motivation}. 
For instance, GeoRSCLIP \citep{zhang2023rs5m} incorporates an additional 5M remote sensing image-text pairs and employs a two-step training process to adapt CLIP to RSITRM. 
This approach undeniably increases the training burden and poses significant challenges for achieving cost-effective performance improvements.

To this end, we introduce a novel framework, iEBAKER, designed to enable a seamless one-step transition from FM to RSITRM, as shown in Fig. \ref{motivation}(b).
Specifically, our approach incorporates an innovative Eliminate Before Align (EBA) strategy with two schemes and a Sort After Reversed Retrieval (SAR) strategy 
to mitigate the adverse effects of weakly correlated pairs. 
Furthermore, we introduce a Keyword Explicit Reasoning (KER) module to enhance the positive role of subtle key concept distinctions. 
We validate the effectiveness of the iEBAKER framework on three widely recognized benchmark datasets, i.e., RSICD \citep{lu2017exploring}, RSITMD \citep{yuan2021exploring}, and NWPU \citep{cheng2022nwpu} datasets. 
Comprehensive experiments reveal that iEBAKER consistently surpasses state-of-the-art methods.

Our contributions are summarized as follows:

\begin{itemize}
  \item To facilitate a one-step transition from  from FM to RSITRM, we propose an Improved Eliminate Before Align strategy with Keyword Explicit Reasoning framework (iEBAKER), 
  which focuses on achieving fine-grained alignment by thoroughly analyzing subtle distinctions and filtering out noise. 
  Unlike the SOTA method, GeoRSCLIP \citep{zhang2023rs5m}, our approach achieves comparable results while utilizing only 4\% of the training data.
  \item To mitigate the negative impact of the weakly correlated pairs, we propose the Eliminate Before Align (EBA) strategy and the Sort After Reversed Retrieval (SAR) strategy.
  The EBA strategy includes two alternative schemes from the perspective of whether local similarity and global similarity affect each other, 
  which both enable autonomously filters out positive sample pairs with low similarities. 
  The SAR strategy employs candidates for reverse retrieval to optimize the results in an offline manner. 
  \item We introduce a Keyword Explicit Reasoning (KER) module, designed to encourage the model to predict subtle distinctions in key concepts within local features of remote sensing images. 
  This module promotes fine-grained contrastive learning, thereby improving the model's ability to differentiate between highly similar sample pairs.
  \item Extensive experiments on three public benchmark datasets, i.e.,  RSICD, RSITMD, and NWPU, showcase that our iEBAKER consistently outperforms
  the state-of-the-arts by a large margin.
\end{itemize}  

It should be noted that this paper extends our conference version \citep{ji2024eliminate} in terms of \textbf{Methodology}, \textbf{Experiments}, and \textbf{Presentation}: 1) we improve our method by: 
i) proposing an alternative EBA strategy by establishing two similarity banks for global and local similarities, respectively, which protects the global similarity from being limited to the threshold selection of local similarity. 
ii) introducing a post-processing method (SAR strategy) to mitigate the negative impact of weakly correlated sample pairs, which has synergistic effect with our proposed EBA strategy from a different perspective.
iii) employing an additional Exponential Moving Average (EMA) training strategy to ensure the model adapt quickly to new patterns while maintaining a balance with historical data,
which provides robustness and enhances overall model performance. This simple yet effective training scheme facilitates the exploitation of the key distinctions among remote sensing text.
2) we conduct additional experiments to demonstrate the effectiveness of the proposed modules, include more recently published works into comparisons, 
   explore the impact of different combinations of training datasets, and conduct qualitative comparison with our previous version \citep{ji2024eliminate}. 
   Extensive experimental results validate that this work achieves much better results than its previous version on all evaluation benchmarks.
3) we include a section on related work according to several relevant aspects of our method, and provide more detailed descriptions about our proposed method for better understanding.

\section{Related work}

\subsection{Remote sensing image-text retrieval}
Remote Sensing Image-Text Retrieval (RSITR), which aims at searching for instances within remote sensing domain from another modality as query, and is initially explored by employing LSTM and various CNN backbones \citep{abdullah2020textrs}. According to the model initialization methods, the existing methods could be roughly categorized into two groups, i.e., traditional methods and Foundation Model (FM) based methods.

Traditional methods randomly initialize the well-designed model, such as utilizing CNNs to extract image features, LSTM or GRU to represent text features, without loading any pre-trained models or parameters, such as \citep{abdullah2020textrs, lv2021fusion, yuan2021exploring, yuan2022mcrn, yuan2022remote, zhang2023hypersphere, pan2023prior, ma2024direction,ji2023knowledge}. In the early stages, research efforts primarily revolve around CNN-based approaches. \citet{abdullah2020textrs} pioneered the exploration of the RSITR problem by employing an average fusion strategy to attain robust representations. \citet{yuan2021exploring} advanced the field further by introducing a visual self-attention module and a fine-grained dataset, i.e., RSITMD.
After that, a large number of studies \citep{yuan2021exploring, yuan2022remote, zhang2023hypersphere, pan2023prior, ma2024direction} focus on refining alignment tailored to the characteristics of RSITR task. For instance, \citet{ji2023knowledge} proposed a knowledge aided learning framework and emphasized the key vocabulary for capturing the subtle differences among images. \citet{zhang2023hypersphere} designed a key-entity attention to keep balance between the visual modality and the textual modality. \citet{pan2023prior} devised a language cycle attention mechanism to address semantic noise issues.  

In recent years, with the flourishing development of FMs \citep{radford2021learning,li2022blip,li2023blip,dai2024instructblip,touvron2023llama,chen2023pali} and their outstanding performance in various downstream tasks, such as image-text retrieval \citep{ji2024hierarchical, zhang2025hierarchical, zhang2024user, zhang2023consensus} and text-based person search \citep{yang2023towards}, researches in RSITR have pivoted towards the transfer from FM to RSITRM \citep{zhang2023rs5m, kuckreja2023geochat}. Specifically, FM based methods resort to the pre-trained models to initialize the well-designed model. For example, \citet{yuan2023parameter} explored multiple Parameter-Efficient Fine-Tuning strategies to transfer the pre-trained CLIP model to the remote sensing domain. \citet{liu2023remoteclip} annotated multiple remote sensing datasets and compared the performance of different large-scale models such as CLIP \citep{radford2021learning}, BLIP \citep{li2022blip}, and ALBEF \citep{li2021align} in the remote sensing domain. \citet{zhang2023rs5m} proposed a 5M remote sensing dataset and achieved excellent performance by employing a two-step approach involving RS pretraining and downstream task fine-tuning to adapt CLIP to the remote sensing domain.

Despite achieving significant advancements in adapting FM to RSITRM, these approaches still require a substantial amount of remote sensing data for pre-training, which inevitably adds the training burden. This work conducts fine-grained alignment through in-depth analysis of subtle distinctions and noise filtration, achieving a one-step training from FM to RSITRM with only relying 4\% of the training data compared with \citep{zhang2023rs5m}.

\subsection{Keyword reasoning in multi-modal learning}
Keyword reasoning aims at taking the advantage of the keywords to reason valuable information in various downstream tasks, such as image-text retrieval \citep{wang2022coder,ji2023knowledge,jiang2023cross,zhang2023consensus}, image captioning \citep{cao2022vision, wang2022multi}, visual grounding \citep{LI2025152,ji2024Progressive}, visual question answering \citep{zhang2022query}, and text-to-image generation \citep{cheng2021rifegan2}.

In the domain of image-text retrieval, \citet{wang2022coder} extracted the keywords informantion as consensus knowledge 
and accounted for statistical co-occurrence correlations among keywords to develop consensus-aware concept representations. 
\citet{zhang2023consensus} constructed a concept graph with high frequency keywords to produce interventional consensus representations, 
thereby uncovering intrinsic associations among concepts. \citet{ji2023knowledge} focused on the keywords within the domain of remote 
sensing and developed a novel framework to learn discriminative representations. 
Jiang and Ye \citep{jiang2023cross} designed an implicit relation reasoning module in a masked language modeling paradigm for excavating the fine-grained
relations between visual and textual tokens.
In the domain of visual question answering, \citet{zhang2022query} 
achieved visual reasoning by utilizing knowledge-augmented queries and memory-augmented attention mechanisms to integrate visual and external knowledge. 
As for image captioning, \citet{cao2022vision} combined memory-based visual representations with consensus knowledge representations to generate image captions.
\citet{zhang2021consensus} proposed to learn consensus representations by aligning visual graphs and textual graphs, and incorporated 
these representations into the grounded captioning pipeline. Different from these existing methods, we propose to facilitate the positive role of 
subtle key concept differences within the limited dataset for achieving a one-step transformation from FM to RSITR task.

\subsection{Learning with noisy correspondence}
Noisy correspondence represents a specific type of labeling error, where mismatched pairs are mistakenly identified as matched pairs.
To reduce the negative impact of the noisy correspondence, numerous methods have been proposed, including the design of robust network architectures, 
the incorporation of regularization, the weighting of different loss terms, and the identification of clean samples \citep{huang2024learning, huang2021learning}.
As a pioneering work, \citet{huang2021learning} proposed a noisy correspondence rectifier for rectifying the matching relationships and achieving robust cross-modal retrieval.
\citet{qin2022deep} enhanced the robustness and reliability of the model by accurately estimating the uncertainty caused by noise.
However, these methods neglect the reliability of the supervision information and cannot guarantee the reliability of the model. 
Subsequently, \citet{hu2023cross} and \citet{yang2023bicro} proposed methods for unbiased estimation and soft label estimation to better reflect the true degree of correspondence.
Although the above sophisticated and well-targeted methods have made great progress, this work addresses the noise correspondence from a novel perspective, i.e., the samples with low similarity are eliminated before alignment during the training.

\section{Method}

\begin{figure*}[t]
  \centering
  \includegraphics[width=\textwidth]{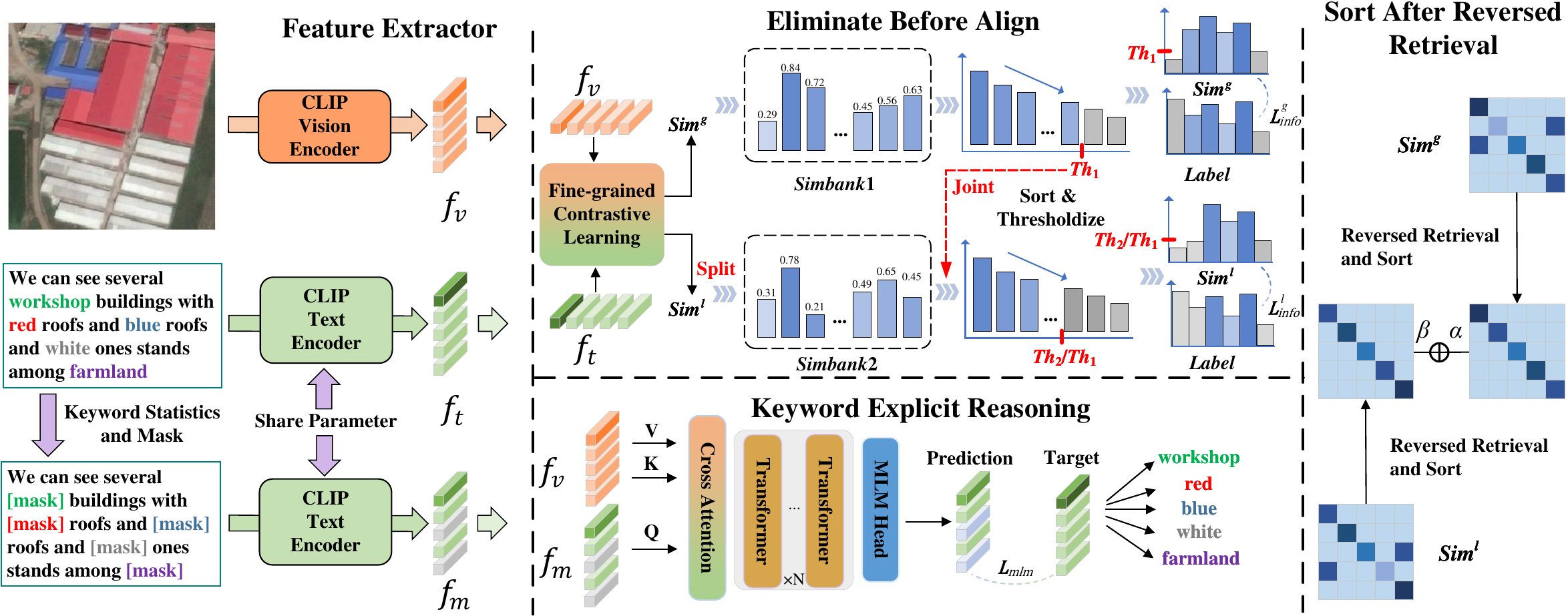} 
  \caption{Overview of our iEBAKER approach, which is composed of four key components: 
  \textbf{A. Feature Extractor:} CLIP~\citep{radford2021learning} is employed as the encoder for both visual and textual modalities. We also conduct word frequency analysis to mask critical keywords. This process yields visual features, textual features, and masked textual features.
  \textbf{B. Eliminate Before Align:} Prior to the alignment step, we eliminate positive sample pairs that exhibit low global similarity, aiming at mitigating the negative impact of the weakly correlated pairs.
  This improved version introduce two specific schemes from the perspective of whether local similarity and global similarity affect each other, i.e., the EBA-Joint and the EBA-Split.
  \textbf{C. Sort After Reversed Retrieval:} A novel post-processing strategy is applied to optimize local and global similarities, respectively.
  \textbf{D. Keyword Explicit Reasoning:} To capture subtle distinctions among remote sensing images, we implement a keyword prediction technique that highlights key concepts, 
  promoting more accurate and fine-grained contrastive learning.
  }
  \label{method}
\end{figure*}

In this section, we present our iEBAKER framework, as depicted in Fig. \ref{method}.
We begin with a detailed introduction of the vision encoder, the text encoder, and the process of keyword statistics and mask generation in Section \ref{feature}. 
Subsequently, we delve into our Eliminate Before Align (EBA) strategy in Section \ref{EBA}, followed by the Sort After Reversed Retrieval (SAR) strategy in Section \ref{EBA}, 
and the Keyword Explicit Reasoning (KER) module in Section \ref{KER}. Lastly, Section \ref{Loss} provides a comprehensive description of the overall loss function and the associated training procedure.

\subsection{Feature extractor}
\label{feature}
\subsubsection{Vision encoder}
Give an input image $I \in R^{(H \times W \times C)}$,  we initially transform $I$ into $N = H \times W/P^{2}$ non-overlapping blocks of fixed size, where $N$ is the number of patches, $H$, $W$, and $C$ represent the height, width, and channel of the image, respectively, $P$ represents the block size.
Subsequently, all blocks are mapped to 1D tokens through a trainable linear projection.
After incorpating positional encoding and an additional $[CLS]$ token, the input block sequence is processed through $L$ layers of transformer blocks to establish the relationship among the blocks.
Finally, all the features undergo linear projection, the embedding of $[CLS]$ token is transformed into the visual global feature $f_{v}^{g}$, and the set $\left\{f_{v}^{1} \ldots f_{v}^{N} \right\}$ represents the visual local features.
The aformentioned process could be simplified as:
\begin{equation}
f_{v}^{g},f_{v}^{1},...,f_{v}^{N} = \varphi(I),
\end{equation}
where $\varphi$ represents vision encoder of CLIP.

\subsubsection{Text encoder}
For a given input text $T$ with $W$ words, we utilize CLIP text encoder to extract representations. Initially, we tokenize the input text by lower-cased Byte Pair Encoding (BPE) with a vocabulary size of 49,152. The text description is surrounded by $[SOS]$ and $[EOS]$ tokens to indicate the start and end of the sequence.
Subsequently, the set $\left\{ {f_{t}^{\text{sos~}},f_{t}^{1}\ldots f_{t}^{\text{eos~}}} \right\}$ is fed to transformer block \citep{vaswani2017attention}, which employs masked self-attention to explore relationships between blocks.
Finally, all the textual features $\left\{ {f_{t}^{\text{sos~}},f_{t}^{1}\ldots f_{t}^{\text{eos~}}} \right\}$ undergo linear projection, 
where $f_{t}^{\text{eos~}}$ is transformed into the textual global feature $f_{t}^{g}$, and the others represent the textual local features.
Similarly, the aformentioned process is simplified as:
\begin{equation}
  f_{t}^{g},f_{t}^{1},...,f_{t}^{W} = \phi(T),
\end{equation}
where $W$ represents the number of words in the input text $T$, $\phi$ represents text encoder of CLIP.

\subsubsection{Keyword statistics and mask generation}\label{KSMG}
We initially perform a statistical analysis to identify key concepts that require masking. Through word frequency analysis across the entire dataset, we exclude common high-frequency words without discriminative value such as ``a'', ``the'', ``of'', etc. 
Subsequently, we select top-$k$ frequency keywords in each dataset, yielding the corresponding keyword list. The process of keyword statistics can be summarized as follows:
\begin{equation}
List_{key} = {Top}_{k}\{Frequency(\sum\nolimits_{i = 1}^{W}T_{i})\}.
\end{equation}
We merge the keyword lists from each datset and remove any duplicate words across them, resulting in the final keyword list for training. 
If a word in the input text $T$ matches the one in the keyword list, it is replaced with ``$[mask]$''.
Accordingly, we generate the masked text $T_{mask}$.
Subsequently, we input the sentences after masking into the text encoder, obtaining the corresponding masked global feature $f_{m}^{g}$ and local features $\left\{ {f_{m}^{1},f_{m}^{2}\ldots f_{m}^{W}} \right\}$ :
\begin{equation}
  f_{m}^{g},f_{m}^{1},...,f_{m}^{W} = \phi(T_{mask}).
\end{equation}

\subsection{Eliminate before align}
\label{EBA}
\begin{figure}[t]
  \centering
  \includegraphics[width=\linewidth]{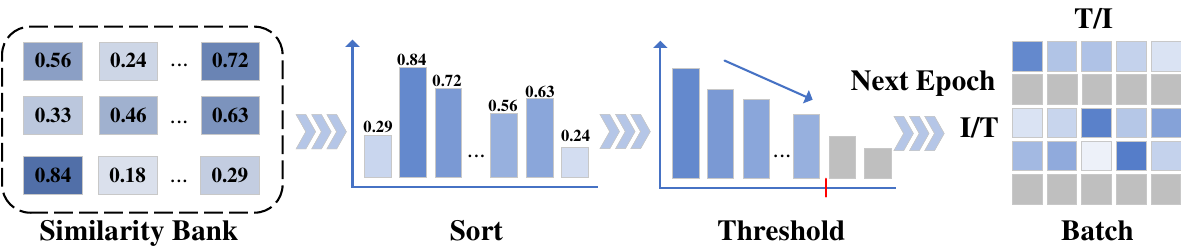} 
  \caption{Explaination of Eliminate Before Align strategy. 
  1) Establish similarity bank to store all global or local similarities on-the-fly. 
  2) Sort all the similarities. 
  3) Set the threshold based on the drop ratio.
  4) Eliminate rows within a batch corresponding to Image-to-Text or Text-to-Image pairs with similarities below the threshold before the alignment in the next epoch.}
  \label{eliminate}
\end{figure}
First, we conduct global alignment between the visual global feature $f_{v}^{g}$ and the textual global feature $f_{t}^{g}$.
Similar to CLIP \citep{radford2021learning}, we compute the global cosine similarity ${Sim}^{g} = cos\left( {f_{v}^{g},f_{t}^{g}} \right)$.

Then, we conduct fine-grained alignment on the local features obtained from the visual and textual encoders. For each input image $I$ and text $T$, 
we obtain visual local features $\left\{f_{v}^{1} \ldots f_{v}^{N} \right\}$ and textual local features $\left\{f_{t}^{1} \ldots f_{t}^{W} \right\}$.
We then compute the local cosine similarity for each local block ${Sim}_{ij}=cos\left( {f_{v}^{i},f_{t}^{j}} \right)$.
Next, we obtain the corresponding local similarities for each image-text pairs by performing two consecutive L2-norm operations on the local blocks:
\begin{equation}
  {Sim}^{l} = \left\| {Sim}_{ij} \right\|_{2,2},
\end{equation}
where $\left\| \cdot  \right\|_{2,2}$ represents the consecutive application of the L2-norm twice.
During the inference stage, we combine the global similarity and local similarity via a weighted approach to obtain the final similarity between image and text:
\begin{equation}
Sim = \alpha{Sim}^{g} + \beta{Sim}^{l},
\label{eq6}
\end{equation}
where $\alpha$ and $\beta$ balance the weight of global similarity and local similarity.

Next, we introduce our proposed EBA strategy as shown in Fig. \ref{eliminate}.
Specifically, we begin with conducting several epochs of regular training without eliminating any training samples, allowing the model to encounter all training samples during this process. 
Subsequently, we establish two Similarity Bank ($SimBank1$ and $SimBank2$), wherein we record the global similarity and local similarity scores of all sample pairs within the current epoch, respectively:
\begin{equation}
  Simbank1 = \{{Sim}_{i}^{g}\}_{i=1}^{L}, Simbank2 = \{{Sim}_{i}^{l}\}_{i=1}^{L},
\end{equation}
where ${Sim}_{i}^{g}$ and ${Sim}_{i}^{l}$ represent the $i$-th global and local similarities, respectively, and $L$ is the total number of image-text pairs in the dataset.
Upon completion of an epoch, we extract all similarity values and sort them in descending order. 
We then select the similarity from the end of the sorted list by the predetermined drop ratio of the data volume, which will be served as the threshold $Th$ for the next epoch:
\begin{align}
  \begin{split}
    Th_1 = Sort\left({Simbank1}\right)\left\lbrack {drop}_{ratio} \right\rbrack,  \\
    Th_2 = Sort\left({Simbank2}\right)\left\lbrack {drop}_{ratio} \right\rbrack,
  \end{split}&
\end{align}
where $Sort$ indicates sorting in descending order and $drop_{ratio}$ represents the specified elimination rate. 
During the training of the next epoch, the global and local similarities of all matched image-text pairs within each batch are compared with $Th_1$ and $Th_2$. 
If the similarity of the current image-text sample pair does not exceed the threshold, its loss is excluded from the current batch. 
Suppose there are $M$ instances within a batch that do not exceed the threshold.
When calculating the loss for image-to-text and text-to-image, the corresponding rows are removed, transforming the $N\times N$ matrix into an $(N-R)\times N$ matrix before alignment:
\begin{align}
  \begin{split}
    &B_{s}^{g} = \left\{ {\sum\limits_{i}^{N}{Sim}_{i}^{g}} \middle| {{Sim}_{i}^{g} > Th_1} \right\} = {\sum\nolimits_{i}^{N - R}{Sim}_{i}^{g}},  \\
    &B_{s}^{l} = \left\{ {\sum\limits_{i}^{N}{Sim}_{i}^{l}} \middle| {{Sim}_{i}^{l} > Th_2} \right\} = {\sum\nolimits_{i}^{N - R}{Sim}_{i}^{l}},
  \end{split}&
\end{align}
where ${Sim}_{i}^{g}$ and ${Sim}_{i}^{l}$ represent the global and local similarities of the $i$-th row, respectively, and $B_{s}^{g}$ and $B_{s}^{l}$ represent the corresponding matrixes within a batch, respectively.

\subsection{Sort after reversed retrieval}\label{SAR}
In this subsection, we perform a post-processing step to optimize the obtained local similarity and global similarity.
The core idea is inspired by \citep{yuan2022remote} and \citep{wang2019matching}, and they argue that an image-text pair must be mutually retrievable.
Thus, we make full use of the top $k$ candidates of local/global similarities and adopt reversed retrieval to optimize the similarities, respectively.
Next, we take global similarity as an example.

First, the top $k$ text candidates and their positions are defined as $t_1$, $t_2$, $\dots$, $t_k$ and $p^t_1$, $p^t_2$, $\dots$, $p^t_k$ given a image query $i$. Second, 
we calculate a optimized similarity $s^g_{i2t}$ by $e^{-\tau(p_m^t+1)}$, where $m \in [1,k]$ and $\tau$ denotes the ranking coefficient. 
Then, we perform the reversed retrieval given the text query $t^m$. Similarly, the top $l$ image candidates and their positions are defined as $i_1$, $i_2$, $\dots$, $i_l$ and $p^i_1$, $p^i_2$, $\dots$, $p^i_l$,
and the optimized similarity $s^g_{t2i}$ by $e^{-\tau(p_n^i+1)}$ is obtained, where $n \in [1,l]$.
Following \citep{yuan2022remote}, we set $s_d$ as the degree of confirmation on similarity predicted by the model, which is calculated as:
\begin{equation}
  s^g_d = \frac{cos(t_m,i_n)}{\sum_{n=0}^{N}cos(t_m,i_n)}.
\end{equation}

Finally, the final similarity based on the local similarity and global similarity is calculated as:
\begin{equation}
  \label{final_sim}
  S = \alpha(s^g_{i2t} + \mu_1s^g_{t2i} + \mu_2s^g_d) + \beta(s^l_{i2t} + \mu_1s^l_{t2i} + \mu_2s^l_d),
\end{equation}
where $\alpha$ and $\beta$ balance the weight of global similarity and local similarity as in Eq. \ref{eq6}, and $\mu_1$ and $\mu_2$ balance the weight of forward retrievable and reversed retrieval.

\subsection{Keyword explicit reasoning}\label{KER}
We first follow \citep{jiang2023cross} and employ a single cross-attention layer, supplemented by several Transformer blocks and a final Masked Language Modeling (MLM) head, to build the keyword reasoning architecture. However, we observe that relying on implicit reasoning over randomly selected tokens as in \citep{jiang2023cross} may overlook subtle yet critical distinctions present in remote sensing images—a concern that will be thoroughly examined in Section \ref{AblationKER}. To address this limitation, we incorporate a Keyword Explicit Reasoning (KER) module, 
which leverages key concepts identified in Section \ref{KSMG}, thus explicitly embedding meaningful keywords into the fine-grained contrastive learning process.

First, we regard the masked textual features $\left\{ {f_{m}^{1}\ldots f_{m}^{W}} \right\}$ $\mathcal{Q}$, the visual features $\left\{ {f_{v}^{g},f_{v}^{1}\ldots f_{v}^{N}} \right\}$ as $\mathcal{K}$ and $\mathcal{V}$.
The function of KER module is to obtain the corresponding predicted probability of the key concepts, which is expressed as follows:
\begin{equation}
    \left\{o_i^m\right\}_{i=1}^{M}=Transformer_{N}\ \text{(}CA\left( \mathcal{Q},\mathcal{K},\mathcal{V} \right)\text{))},
\end{equation}
where $CA$ represents the defined cross attention layer for reasoning the relationship among $\mathcal{Q}$, $\mathcal{K}$, and $\mathcal{V}$, 
$Transformer_N$ represents $N$ Transformer blocks. 
To obtain the corresponding predicted probability, we further insert a MLP architecture (dubbed as $MLM_{head}$), comprising a linear layer, a QuickGELU activation layer, a LayerNorm layer, and an additional linear layer,
which is defined as follows:
\begin{equation}
  \left\{ p_{i}^{m} |m \in List_{key}\right\}_{i = 1}^{M}=MLM_{head}(o_i^m),
\end{equation}
where $\left\{ p_{i}^{m} |m \in List_{key}\right\}_{i = 1}^{M}$ denotes the predicted probability $p$ at position $i$ for the mask $m$ of the $List_{key}$.
The loss function is defined as follows:
\begin{equation}
  \mathcal{L}_{mlm} = - \frac{1}{MV}{\sum\limits_{m = 1}^{M}{{\sum\limits_{i = 1}^{V}{y_{i}^{m}{\log\frac{\exp\left( p_{i}^{m} \right)}{{\sum\limits_{j = 1}^{V}\,}\,{\exp\left( p_{j}^{m} \right)}}}}},}}
\end{equation}
where $M$ represents the number of masked tokens, $V$ is the vocabulary size of CLIP, $y_{i}^{m}$ is the one-hot distribution of the $m$-th masked word corresponding to the $i$-th token.

\subsection{Loss function and training process}\label{Loss}
To achieve fine-grained alignment, we utilize the widely adopted InfoNCE loss function \citep{oord2018representation}, applying it to both global and local similarity measures. 
This can be mathematically represented as follows:
\begin{equation} 
  \mathcal{L}_{info} = - \frac{1}{N}{\sum\limits_{j = 1}^{N}\left( {log\frac{exp\left( {s^{vt^+}_{j}/\gamma} \right)}{\sum_{i = 1}^{N}\,exp\left( {s_{ij}^{vt}/\gamma} \right)}
                      - log\frac{exp\left( {s^{tv^+}_{j}/\gamma} \right)}{\sum_{i = 1}^{N}\,exp\left( {s_{ij}^{tv}/\gamma} \right)}} \right)}, 
\end{equation}
\noindent where $s^{vt^+}$ and $s^{tv^+}$ represent the positive pairs, $\sum_{i = 1}^{N}s_{ij}^{vt}$ and $\sum_{i = 1}^{N}s_{ij}^{tv}$ respectively represent the sum of each row in the similarity matrices for Image-to-Text or Text-to-Image alignments, $\gamma$ represents the temperature hyper-parameter, $N$ represents the batch size.
For the matrices corrected by the EBA strategy during training, we eliminate the corresponding noisy image-text pairs and make the following adjustments to the InfoNCE loss:
\begin{equation}
  \mathcal{\widetilde{L}}_{info} = - \frac{1}{N}{\sum\limits_{j = 1}^{N}\left( {log\frac{exp\left( {s^{vt^+}_{j}/\gamma} \right)}{\sum_{i = 1}^{N-R}\,exp\left( {s_{ij}^{vt}/\gamma} \right)} - log\frac{exp\left( {s^{tv^+}_{j}/\gamma} \right)}{\sum_{i = 1}^{N-R}\,exp\left( {s_{ij}^{tv}/\gamma} \right)}} \right)},
\end{equation} 
where $\sum_{i = 1}^{N-R}s_{ij}^{vt}$ and $\sum_{i = 1}^{N-R}s_{ij}^{tv}$ respectively represent the sum of each row in the similarity matrices for Image-to-Text or Text-to-Image after removing $R$ rows.

Both global and local alignment utilize the InfoNCE loss, while the modeling of masked attributes employs the MLM loss. 
We set a drop epoch $K$, during which the original InfoNCE loss is applied, allowing the model full exposure to the entire dataset. 
After surpassing epoch $K$, we transition to a modified version of the InfoNCE loss to filter out noise and focus on more relevant data. The overall loss function is expressed as follows:
\begin{equation}
  \mathcal{L}_{total}=\begin{cases}
    {\mathcal{~}\mathcal{L}}_{info}^{g} + {\mathcal{~}\mathcal{L}}_{info}^{l} + \gamma\mathcal{L}_{mlm}, \quad if \ epoch <  K, \\
    {\mathcal{~}\mathcal{\widetilde{L}}}_{info}^{g} + {\mathcal{~}\mathcal{\widetilde{L}}}_{info}^{l} + \gamma\mathcal{L}_{mlm}, \quad if \ epoch \geq K, 
  \end{cases} \label{eq15}
\end{equation}

As we all know, remote sensing images have the characteristics of larger intra-class variance and a smaller inter-class variance \citep{ji2023dual}.
In other words, the obvious difference of keywords among samples is not conducive to the training of the model.
Thus, we employ Exponential Moving Average (EMA) \citep{tarvainen2017mean} to train the entire network, which significantly enhances the ability to exploit the critical distinctions 
found within remote sensing text by carefully identifying and leveraging these key differences. 
This technique allows us to smooth the learning process by updating the model parameters more gradually. 
\section{Experiments}

\subsection{Datasets and settings}
\subsubsection{Datasets}
In our experiments, we employ three benchmark datasets, RSICD \citep{lu2017exploring}, RSITMD \citep{yuan2021exploring}, and NWPU \citep{cheng2022nwpu}, to validate the effectiveness of our approach. 
Adhering to the methodology of RemoteCLIP \citep{liu2023remoteclip}, we compute $p$-Hash values for image-text pairs and set a threshold of 2 to merge these three datasets, thereby eliminating redundant images. 
The RSICD dataset, which is the most widely used in the context of remote sensing image-text retrieval (RSITR), contains 10,921 images, 
each with dimensions of 224$\times$224 pixels. The RSITMD dataset consists of 4,743 images, with each image measuring 256$\times$256 pixels, while the NWPU dataset includes 31,500 images, also sized at 256$\times$256 pixels. 
Following the protocol established in \citep{yuan2021exploring}, we divide these three datasets into train sets (80\%), validation sets (10\%), and test sets (10\%).

\subsubsection{Evaluation metrics}
We employ the popular Recall at $k$ ($R@k$, $k$=1,5,10) and mean Recall (mR) as the evaluation metrics for Caption Retrieval (Image-to-Text) and Image Retrieval (Text-to-Image). Specifically, $R@k$ measures the percentage of ground truth instances within the top $k$ samples, offering a measure of precision at different levels.
mR calculates the average value across all six $R@k$ metrics, thereby delivering a comprehensive evaluation of the overall performance. 

\subsection{Implementation details}
The iEBAKER framework is implemented using the RemoteCLIP codebase \citep{liu2023remoteclip} and our previous version \citep{ji2024eliminate}, with the ViT-B-32 architecture 
provided by OpenCLIP \citep{cherti2023reproducible}. 
We train the model for 10 epochs with a batch size of 100, employing the Adam optimizer \citep{kingma2014adam}. A linear warm-up followed by a cosine learning rate scheduler is utilized, 
with the learning rate set to 1.5e-5 and weight decay to 0.7. The warm-up period is configured for 200 iterations, and the maximum gradient norm is set to 50. 
For the EBA strategy, the drop epoch $K$ is set to 4 with a drop ratio of 1\%. 
The KER transformer block count $N$ is set to 4, and the ranking coefficient $\tau$ in Section \ref{SAR} is set to 0.05. 
For the final similarity as described in Eq. \ref{final_sim}, we follow \citep{yuan2022remote}, and set $\mu_1=0.5$ and $\mu_2=1.25$. 
All the experiments are implemented with PyTorch and trained with a single NVIDIA GeForce RTX 4090 GPU.

\setlength{\tabcolsep}{3pt}
\begin{table*}[!t]
  \centering
  \caption{Comparison results of the Caption Retrieval and Image Retrieval on \textbf{RSICD} dataset. The best and second-best results are hightlighted in bold and underlined.}
  \renewcommand\arraystretch{1.1}
  \begin{tabular}{lccccccccc}
  \toprule
  \multirow{2}{*}{Approach} & \multirow{2}{*}{Backbone} & \multicolumn{3}{c}{Caption Retrieval} & \multicolumn{3}{c}{Image Retrieval} & \multirow{2}{*}{mR} \\
                            &                & R@1           & R@5           & R@10          & R@1           & R@5           & R@10          &                     \\
  \midrule
  VSE++$_{\rm_{BMVC'18}}$\citep{faghrivse++}     & ResNet18/Bi-GRU             & 3.38          & 9.51          & 17.46         & 2.82          & 11.32         & 18.10         & 10.43               \\
  LW-MCR$_{\rm_{TGRS'21}}$\citep{yuan2021lightweight}              & ResNet18/Bi-GRU       & 3.29          & 12.52         & 19.93         & 4.66          & 17.51         & 30.02         & 14.66               \\
  AMFMN$_{\rm_{TGRS'22}}$\citep{yuan2021exploring}                 & ResNet18/Bi-GRU                                                             & 5.39          & 15.08         & 23.40         & 4.90          & 18.28         & 31.44         & 16.42               \\
  GaLR$_{\rm_{TGRS'22}}$\citep{yuan2022remote}                     & ResNet18/Bi-GRU                 & 6.59          & 19.85         & 31.04         & 4.69          & 19.48         & 32.13         & 18.96               \\
  Multilanguage$_{\rm_{JSTARS'22}}$\citep{al2022multilanguage}     & ViT-B-32/BERT       & 10.70         & 29.64         & 41.53         & 9.14          & 28.96         & 44.59         & 27.42               \\
  SWAN$_{\rm_{ICMR'23}}$\citep{pan2023reducing}                    & ResNet50/Glove     & 7.41          & 20.13         & 30.86         & 5.56          & 22.26         & 37.41         & 20.61               \\
  PIR$_{\rm_{ACMMM'23}}$\citep{pan2023prior}                       & Swin-T/BERT               & 9.88          & 27.26         & 39.16         & 6.97          & 24.56         & 38.92         & 24.46               \\
  KAMCL$_{\rm_{TGRS'23}}$\citep{ji2023knowledge}                   & ResNet101/Bi-GRU                                                           & 12.08         & 27.26         & 38.70         & 8.65          & 27.43         & 42.51         & 26.10               \\
  PE-RSITR$_{\rm_{TGRS'23}}$\citep{yuan2023parameter}              & CLIP(ViT-B-32)             & 14.13         & 31.51         & 44.78         & 11.63         & 33.92         & 50.73         & 31.12               \\
  MSA$\rm_{TGRS'24}$\citep{transcending} & CLIP-RN50/BERT & 10.16 & 25.71 & 36.96 & 7.87 & 25.67 & 41.85 & 24.70  \\
  RemoteCLIP$_{\rm_{TGRS'24}}$\citep{liu2023remoteclip}           & CLIP(ViT-B-32)               & 17.02         & 37.97         & 51.51         & 13.71         & 37.11         & 54.25         & 35.26               \\
  GeoRSCLIP$_{\rm_{TGRS'24}}$\citep{zhang2023rs5m}                & CLIP(ViT-B-32)           & 21.13         & 41.72        & 55.63         & 15.59         & 41.19         & 57.99         & 38.87               \\
  AIR$\rm_{ACMMM'24}$\citep{yang2024accurate}                      & CLIP(ViT-B-32)                                                                & 18.85         & 39.07        & 51.78         & 14.24         & 39.03         & 54.49         & 36.24  \\
  UrbanCross$\rm_{ACMMM'24}$\citep{zhong2024urbancross}            & ViT-L-14/Transformer     & 17.52         & 38.49        & 51.86         & 14.52         & 40.89         & 57.67         & 36.83 \\ 
  SWAP$\rm_{JSTARS'25}$ \citep{10855571} & RemoteCLIP & 18.66 & 39.52 & 53.61 & 15.33 & 40.86 & 57.73 & 37.62 \\
  EBAKER$_{\rm_{ACMMM'24}}$\citep{ji2024eliminate}                 & CLIP(ViT-B-32)            & {21.87}         & {44.46}          & {58.92}         & {17.37}        & {43.00}         & {58.55}        & {40.70}               \\
  iEBAKER-Joint(Ours)   & CLIP(ViT-B-32)   & \underline{24.25}   & \underline{49.41}    & \underline{63.68}   & \underline{18.94}   & \underline{45.56}   & \underline{63.40}    & \underline{44.21}               \\   
  iEBAKER-Split(Ours)  & CLIP(ViT-B-32)        & \textbf{25.80}      & \textbf{50.32}       & \textbf{64.04}      & \textbf{19.76}      & \textbf{47.06}      & \textbf{63.42}    & \textbf{45.07}               \\          
  \bottomrule
  \end{tabular}
  \label{SOTA_1}
  \end{table*}

\subsection{Comparisons with the SOTA methods}
In this section, we present the quantitative results of our approach compared to state-of-the-art methods on three public benchmark datasets. 
We categorize existing methods into two groups: traditional methods and Foundation Model (FM) based methods.
The traditional methods considered include VSE++ \citep{faghrivse++}, LW-MCR \citep{yuan2021lightweight}, AMFMN \citep{yuan2021exploring}, GaLR \citep{yuan2022remote},  
Multilanguage \citep{al2022multilanguage}, SWAN \citep{pan2023reducing}, PIR \citep{pan2023prior}, and KAMCL \citep{ji2023knowledge}.
For FM based methods, the involved methods includes PE-RSITR \citep{yuan2023parameter}, MSA \citep{transcending}, RemoteCLIP \citep{liu2023remoteclip}, GeoRSCLIP \citep{zhang2023rs5m}, AIR \citep{yang2024accurate}, UrbanCross \citep{zhong2024urbancross}, SWAP\citep{10855571},
and our previous version \citep{ji2024eliminate}. 
Our proposed iEBAKER falls within the category of FM based methods.
The experimental results for the RSICD \citep{lu2017exploring}, RSITMD \citep{yuan2021exploring}, and NWPU \citep{cheng2022nwpu} datasets are summarized in Tables \ref{SOTA_1}--\ref{SOTA_3}. 
In addition to performance metrics, we provide the vision and text backbone architectures (denoted as "vision/text"), as well as the total training and test set sizes for each method. 

Note that different methods employ different training datasets. Specifically, RemoteCLIP\citep{liu2023remoteclip} and GeoRSCLIP\citep{zhang2023rs5m} collect 0.87M and 5.07M remote sensing data for training, respectively. Differently, we combine the training sets of RSICD, RSITMD, and NWPU for training, only 0.2M. The other involved comparison methods leverage the training sets of the respective datasets for training, and evaluate the performance on the corresponding test sets. In Tables \ref{SOTA_1}--\ref{SOTA_3}, ``Joint'' refers to our previous version of the Eliminate Before Align (EBA) strategy, while ``Split'' corresponds to this improved version. Based on these results, we derive the following key observations and conclusions.

\subsubsection{Quantitative comparison on RSICD, RSITMD, and NWPU datasets}
For the \textbf{RSICD} dataset, our EBAKER, iEBAKER-Joint, and iEBAKER-Split methods notably outperforms all competing methods across a range of evaluation metrics. 
Compared with GeoRSCLIP \citep{zhang2023rs5m}, the existing best method of FM based method, our iEBAKER-Joint, and iEBAKER-Split methods surpasses it on all evaluation metrics, while utilizing only 0.2 million data samples--just 4\% of the 5.07 million samples used by GeoRSCLIP. Particularly noteworthy is the 3.29\%, 8.05\%, and 8.41\% improvement in caption retrieval R@10, 1.78\%, 3.35\%, and 4.17\% enhancement in image retrieval R@1, and overall mR increase of 1.83\%, 5.34\%, and 6.20\% 
for EBAKER, iEBAKER-Joint, and iEBAKER-Split, respectively. For the \textbf{RSITMD} dataset, our EBAKER, iEBAKER-Joint, and iEBAKER-Split methods achieve 1.77\%, 2.88\%, and 3.76\% improvement in caption retrieval R@1, 3.01\%, 2.57\%, and 2.48\% 
enhancement in image retrieval R@1, and overall mR increase of 1.51\%, 3.65\%, and 3.74\%, respectively. These findings further emphasize the comprehensive performance superiority of our approach over GeoRSCLIP. Following \citep{ji2023knowledge}, we also conduct comparative experiments on the NWPU dataset, where the results of RemoteCLIP \citep{liu2023remoteclip} are reproduced by fine-tuning the pre-trained weight matrixes provided in \citep{yuan2022remote} on the NWPU dataset. Specifically, our observations reveal a 3.49\%, 5.71\%, and 5.14\% improvement in caption retrieval R@10, a 1.02\%, 2.14\%, and 1.97\% enhancement in image retrieval R@10, and an overall mR increase of 1.46\%, 2.64\%, and 2.86\% for EBAKER, iEBAKER-Joint, and iEBAKER-Split, respectively. These results clearly demonstrate that our methods not only achieve superior performance but also do so with significantly less data.

\setlength{\tabcolsep}{3pt}
\begin{table*}[!t]
  \centering
  \caption{Comparison results of the caption retrieval and image retrieval on \textbf{RSITMD} datasets. The best and second-best results are hightlighted in bold and underlined.}
  \renewcommand\arraystretch{1.1}
  \begin{tabular}{lccccccccc}
  \toprule
  \multirow{2}{*}{Approach} & \multirow{2}{*}{Backbone} & \multicolumn{3}{c}{Caption Retrieval} & \multicolumn{3}{c}{Image Retrieval} & \multirow{2}{*}{mR} \\
                            &                & R@1           & R@5           & R@10          & R@1           & R@5           & R@10          &                     \\
  \midrule
  VSE++$_{\rm_{BMVC'18}}$\citep{faghrivse++}                       & ResNet18/Bi-GRU           & 10.38         & 27.65         & 39.60          & 7.79          & 24.87         & 38.67         & 24.83               \\
  LW-MCR$_{\rm_{TGRS'21}}$\citep{yuan2021lightweight}              & ResNet18/Bi-GRU                     & 10.18         & 28.98         & 39.82         & 7.79          & 30.18         & 49.78         & 27.79               \\
  AMFMN$_{\rm_{TGRS'22}}$\citep{yuan2021exploring}                & ResNet18/Bi-GRU           & 11.06         & 29.20         & 38.72         & 9.96          & 34.03         & 52.96         & 29.32               \\
  GaLR$_{\rm_{TGRS'22}}$\citep{yuan2022remote}                    & ResNet18/Bi-GRU            & 14.82         & 31.64         & 42.48         & 11.15         & 36.68         & 51.68         & 31.41               \\
  Multilanguage$_{\rm_{JSTARS'22}}$\citep{al2022multilanguage}     & ViT-B-32/BERT        & 19.69         & 40.26         & 54.42         & 17.61         & 49.73         & 66.59         & 41.38               \\
  SWAN$_{\rm_{ICMR'23}}$\citep{pan2023reducing}                  & ResNet50/Glove       & 13.35         & 32.15         & 46.90         & 11.24         & 40.40         & 60.60         & 34.11               \\
  PIR$_{\rm_{ACMMM'23}}$\citep{pan2023prior}                    & Swin-T/BERT       & 18.14         & 41.15         & 52.88         & 12.17         & 41.68         & 63.41         & 38.24               \\
  KAMCL$_{\rm_{TGRS'23}}$\citep{ji2023knowledge}                 & ResNet101/Bi-GRU         & 16.51         & 36.28         & 49.12         & 13.50         & 42.15         & 59.32         & 36.14               \\
  PE-RSITR$_{\rm_{TGRS'23}}$\citep{yuan2023parameter}             & CLIP(ViT-B-32)      & 23.67         & 44.07         & 60.36         & 20.10         & 50.63         & 67.97         & 44.47               \\
  MSA$\rm_{TGRS'24}$\citep{transcending} & CLIP-RN50/BERT  & 22.35 & 42.92 & 55.75 & 15.18 & 47.35 & 64.73 & 41.38 \\
  RemoteCLIP$_{\rm_{TGRS'24}}$\citep{liu2023remoteclip}          & CLIP(ViT-B-32)        & 27.88         & 50.66         & 65.71         & 22.17         & 56.46         & 73.41         & 49.38               \\
  GeoRSCLIP$_{\rm_{TGRS'24}}$\citep{zhang2023rs5m}               & CLIP(ViT-B-32)                      & 32.30        & 53.32         & 67.92         & 25.04         & 57.88         & 74.38         & 51.81               \\
  AIR$\rm_{ACMMM'24}$\citep{yang2024accurate}                      & CLIP(ViT-B-32)             & 29.20         & 49.78         & 65.27         & 26.06         & 57.04         & 73.98         & 50.22 \\
  UrbanCross$\rm_{ACMMM'24}$\citep{zhong2024urbancross}            & ViT-L-14/Transformer          & 27.98          & 51.68        & 65.56         & 23.66         & 58.44         & 73.78         & 50.18 \\
   SWAP$\rm_{JSTARS'25}$ \citep{10855571} & RemoteCLIP & 27.88 & 51.76 & 64.82 & 25.27 & 58.23 & 75.27 & 50.54 \\
  EBAKER$_{\rm_{ACMMM'24}}$\citep{ji2024eliminate}                 & CLIP(ViT-B-32)      & {34.07}         & {54.20}         & {67.95}         & \textbf{28.05}         & {60.35}         & {75.31}         & {53.32}    \\
  iEBAKER-Joint(Ours)     & CLIP(ViT-B-32)   & \underline{35.18}  & \underline{57.30}         & \textbf{71.90}     &  \underline{27.61}  & \textbf{62.43}         & \underline{78.36}   & \underline{55.46}                 \\
  iEBAKER-Split(Ours)     & CLIP(ViT-B-32)      & \textbf{36.06}     & \textbf{58.63}     & \underline{71.68}         &  {27.52}        & \underline{60.84}         & \textbf{78.58}    & \textbf{55.55}                 \\
  \bottomrule
  \end{tabular}
  \label{SOTA_2}
  \end{table*}

\setlength{\tabcolsep}{3pt}
\begin{table*}[!t]
  \centering
  \caption{Comparison results of the caption retrieval and image retrieval on \textbf{NWPU} datasets. The best and second-best results are hightlighted in bold and underlined.}
  \renewcommand\arraystretch{1.1}
  \begin{tabular}{lccccccccc}
  \toprule
  \multirow{2}{*}{Approach} & \multirow{2}{*}{Backbone} & \multicolumn{3}{c}{Caption Retrieval} & \multicolumn{3}{c}{Image Retrieval} & \multirow{2}{*}{mR} \\
                            &                & R@1           & R@5           & R@10          & R@1           & R@5           & R@10          &                     \\
  \midrule
  VSE++$_{\rm_{BMVC'18}}$\citep{faghrivse++}                         & ResNet18/Bi-GRU        & 4.84          & 12.89         & 20.94         & 4.38          & 13.61         & 24.12         & 13.46               \\
  AMFMN$_{\rm_{TGRS'22}}$\citep{yuan2021exploring}                   & ResNet18/Bi-GRU           & 11.49         & 38.75         & 57.73         & 8.63          & 30.25         & 46.48         & 32.22               \\
  KAMCL$_{\rm_{TGRS'23}}$\citep{ji2023knowledge}                     & ResNet101/Bi-GRU      & 21.02         & 57.33         & 74.41         & 12.74         & 38.03         & 53.90         & 42.90               \\
  RemoteCLIP$_{\rm_{TGRS'24}}$\citep{liu2023remoteclip}              & CLIP(ViT-B-32)      & 24.57         & 57.75         & 74.19         & 14.95         & 40.17        & 55.75         & 44.56               \\
  EBAKER$_{\rm_{ACMMM'24}}$\citep{ji2024eliminate}                   & CLIP(ViT-B-32)          & {24.98}        & {60.95}         & {77.68}         & 14.55         & {41.16}         & {56.77}         & {46.02}              \\
  iEBAKER-Joint (Ours)        & CLIP(ViT-B-32)    & \underline{26.51}      & \underline{61.49}          & \textbf{79.90}            & \textbf{15.42} & \textbf{41.99}           & \textbf{57.89}         & \underline{47.20} \\
  iEBAKER-Split (Ours)     & CLIP(ViT-B-32)     & \textbf{26.79}           & \textbf{63.71}          & \underline{79.33}            & \underline{15.02} & \underline{41.91}   & \underline{57.72}    & \textbf{47.42} \\
  \bottomrule
  \end{tabular}
  \label{SOTA_3}
  \end{table*}
  
\setlength{\tabcolsep}{3pt}
  \begin{table*}[!t]
    \centering
    \caption{Ablation experiments with different modules of on RSICD Test Set. The ``EBA-J'' refers to our previous version of the EBA strategy, and ``EBA-S'' corresponds to this improved version.}
    \begin{tabular}{cccccccccccccc}
    \toprule
    \multirow{2}{*}{Method} &  \multicolumn{6}{c}{Modules/Strategies} & \multicolumn{3}{c}{Caption Retrieval}    & \multicolumn{3}{c}{Image Retrieval}    & \multirow{2}{*}{mR} \\
                &                Local & KER        & EBA-J      & EBA-S      & EMA         & SAR    & R@1            & R@5            & R@10           & R@1            & R@5            & R@10           &                     \\
    \midrule
    1           &                      &            &            &            &             &        & 18.85          & 39.35          & 54.08          & 15.93          & 41.45          & 57.85          & 37.92\\
    2           &            \checkmark&            &            &            &             &        & 20.95          & 42.45          & 56.45          & 16.51          & 41.99          & 57.77          & 39.35\\
    3           &                      & \checkmark &            &            &             &        & 19.58          & 42.63          & 56.72          & 16.61          & 42.20          & 58.23          & 39.33\\
    4           &                      &            &  \checkmark&            &             &        & 19.58          & 41.72          & 56.36          & 16.98          & 43.18          & 58.19          & 39.34\\
    5           &            \checkmark& \checkmark &            &            &             &        & 21.23          & 43.92          & 58.37          & 17.18          & 43.22          & 58.79          & 40.45\\
    6           &            \checkmark&            &  \checkmark&            &             &        & 20.59          & 44.65          & 57.64          & 17.24          & 42.29          & 57.93          & 40.05\\
    7           &                      & \checkmark &  \checkmark&            &             &        & 20.85          & 42.99          & 56.81          & 17.71          & 43.27          & 57.94          & 39.93\\
    8           &            \checkmark& \checkmark &  \checkmark&            &             &        & 21.87          & 44.46          & 58.92          & 17.37          & 43.00          & 58.55          & 40.70\\
    9           &            \checkmark& \checkmark &  \checkmark&            &  \checkmark&        & 22.96          & 44.46          & 59.65          & 19.30          & 45.34          & 62.05          & 42.30\\
    10           &            \checkmark& \checkmark &            &  \checkmark&  \checkmark&        & 23.51          & 45.65          & 59.65          & 19.30          & 45.89          & 61.77          & 42.63\\
    11          &            \checkmark& \checkmark &  \checkmark&            &  \checkmark & \checkmark& {24.25} & {49.41} & {63.68} & {18.94} & {45.56} & {63.40} & {44.21}\\
    12          &            \checkmark& \checkmark &            &  \checkmark&  \checkmark &  \checkmark & \textbf{25.80}     & \textbf{50.32}          & \textbf{64.04}          & \textbf{19.76}          & \textbf{47.06}          & \textbf{63.42}          & \textbf{45.07}\\
    \bottomrule
    \end{tabular}
    \label{structure}
  \end{table*}

\subsubsection{Comparison between traditional and FM based methods} 
Compared with the traditional approaches, FM based methods generally deliver superior performance through fine-tuning. 
However, they typically require more training data. For instance, GeoRSCLIP\citep{zhang2023rs5m} necessitates an additional 5M remote sensing dataset for its 
RS pretraining process, as shown in Fig. \ref{motivation} (a). Our methods—EBAKER, iEBAKER-Split, and iEBAKER-Joint—achieve a commendable balance between performance 
and computational efficiency by relying solely on the RSICD, RSITMD, and NWPU datasets, thus obviating the need for additional remote sensing data for pretraining.
Compared with KAMCL \citep{ji2023knowledge}, the leading traditional method, our EBAKER demonstrates notable performance enhancements of 14.60\% and 17.18\% in 
mR on the RSICD and RSITMD datasets, respectively. The iEBAKER-Joint shows even greater improvements of 18.11\% and 19.32\%, while iEBAKER-Split 
achieves 18.97\% and 19.41\% enhancements, respectively. Furthermore, in contrast to GeoRSCLIP \citep{zhang2023rs5m}, our methods achieve competitive performance 
improvements with a streamlined, one-step fine-tuning process, thereby eliminating the need for additional pretraining samples.

\subsubsection{Comparison among EBAKER, iEBAKER-Joint, and iEBAKER-Split}
As shwon in Tables \ref{SOTA_1}--\ref{SOTA_3}, our iEBAKER-Joint and iEBAKER-Split exhibit superior performance with the same train sets and backbones compared with our previous version \citep{ji2024eliminate}.
For instance, iEBAKER-Joint improves the mR metric by 3.51\%, while iEBAKER-Split achieves a 4.37\% improvement. 
In terms of caption retrieval, iEBAKER-Joint and iEBAKER-Split increase the R@1 score by 2.38\% and 3.97\%, respectively. 
For image retrieval, they further enhance R@1 by 1.57\% and 2.39\%, respectively. 
These results underscore the enhanced retrieval capabilities of our iEBAKER variants over earlier models.

\subsection{Ablation studies}
In this section, we conduct a series of ablation studies to assess the performance contributions of individual modules, and examine the effects of various training configurations. 
These experiments aim to isolate the impact of each component on the overall model performance, providing deeper insights into the effectiveness of our framework and the influence of specific design choices.
Unless otherwise specified, the EBA-Joint strategy is selected in all ablation studies.

\subsubsection{Different configurations of the iEBAKER framework}
To begin with, we first validate the effectiveness of different modules in our iEBAKER framework, as shown in Table \ref{structure}. 
Experimentally, we choose the original CLIP \citep{radford2021learning} as the baseline, and incorporate the local alignment (Local), the KER, the EBA-Joint (EBA-J), the EBA-Split (EBA-S), 
the EMA, and the SAR, respectively. 
Compared with the baseline (Method 1), Methods 2-4 result in respective improvements of 1.43\%, 1.41\%, 1.42\% in terms of mR. 
Subsequently, we conduct the combinations of each two modules and find that local alignment with KER module yields promising results, with an improvement of approximately 1.10\% compared with utilizing either mechanism individually. 
This may be attributed to the fact that the reasoning ability of KER explicitly manifests in the fine-grained local alignment.
The effectiveness of EBA-S could be demonstrated by Methods 9 and 10, and Methods 11 and 12, respectively.
Based on these, the integrations of EMA and SAR in Methods 9 and 11 substantially bring about 1.60\% and 1.91\% improvements on mR, respectively.

\setlength{\tabcolsep}{7pt}
\begin{table}[t!]
  \centering
  \caption{Ablation on the ratio of global and local alignment on RSICD Test Set.}
  \resizebox{\linewidth}{!}{
  \begin{tabular}{cccccccccc}
  \toprule
  \multirow{2}{*}{Method} & \multirow{2}{*}{$\alpha$} & \multirow{2}{*}{$\beta$} & \multicolumn{3}{c}{Caption Retrieval}    & \multicolumn{3}{c}{Image Retrieval} & \multirow{2}{*}{mR} \\
                       &                         &                        & R@1            & R@5            & R@10           & R@1            & R@5         & R@10           &                     \\
  \midrule
  1                    & 1                       & 0                      & 22.32          & 44.10          & 58.01          & 16.07          & 42.07       & 58.06          & 40.10                \\
  2                    & 0.9                     & 0.1                    & \textbf{22.96} & 43.82          & 58.37          & 16.45          & 42.29       & 58.50           & 40.40                \\
  3                    & 0.8                     & 0.2                    & 22.78          & 44.46          & 58.37          & 16.93          & 42.36       & 58.48          & 40.56               \\
  4                    & 0.7                     & 0.3                    & 22.14          & 44.28          & 58.28          & 17.20           & 42.84       & \textbf{59.65} & 40.56               \\
  5                    & 0.6                     & 0.4                    & 21.87          & 44.46          & \textbf{58.92} & 17.37          & \textbf{43.00} & 58.55          & \textbf{40.70}       \\
  6                    & 0.5                     & 0.5                    & 21.87          & \textbf{44.74} & 58.65          & 17.35          & 42.80        & 58.41          & 40.63               \\
  7                    & 0.4                     & 0.6                    & 21.32          & 44.65          & 58.83          & \textbf{17.47} & 42.73       & 58.46          & 40.58               \\
  8                    & 0.3                     & 0.7                    & 21.04          & 44.28          & 58.74          & 17.42          & 42.63       & 58.39          & 40.42               \\
  9                    & 0.2                     & 0.8                    & 21.04          & 43.92          & 58.37          & 17.24          & 42.58       & 58.33          & 40.25               \\
  10                   & 0.1                     & 0.9                    & 21.13          & 43.73          & 58.28          & 17.13          & 42.62       & 58.12          & 40.17               \\
  11                   & 0                       & 1                      & 21.23          & 43.82          & 58.28          & 17.13          & 42.62       & 58.04          & 40.19              \\
  \bottomrule
  \end{tabular}}
  \label{global-local}
\end{table}

\subsubsection{Impact of the ratio of global and local alignment}
We vary the weights assigned to the global and local alignments to further investigate their impact, i.e., $\alpha$ and $\beta$.
The results are shown in Table \ref{global-local}, where the sum of the weights for global and local alignment is maintained to 1. 
From the results, the optimal balance is achieved with weights of 0.6 for global alignment and 0.4 for local alignment. 
In this configuration, the mR reaches 40.70\%, which represents a 0.60\% improvement over using only global features and a 0.51\% improvement over relying solely on local features. 
These findings demonstrate that global and local information complement each other, enabling fine-grained contrastive learning to better capture and distinguish intricate details within remote sensing images. 
It should be note that we do not introduce SAR and EMA in this ablation study for fair comparison.

\subsubsection{Impact of different mask strategies}
\label{AblationKER}
\begin{table}[!t]
  \centering
  \caption{Impact of different mask strategies on RSICD Test Set.}
  \resizebox{\linewidth}{!}{
  \begin{tabular}{lccccccc}
  \toprule
  \multirow{2}{*}{Approach} & \multicolumn{3}{c}{Captiion Retrieval}    & \multicolumn{3}{c}{Image Retrieval}   & \multirow{2}{*}{mR} \\
                            & R@1            & R@5            & R@10           & R@1           & R@5            & R@10           &                     \\
  \midrule
  IRR\citep{jiang2023cross}  & 20.04          & 41.81          & 55.35          & 16.93         & 42.93 & 58.30           & 39.23               \\
  MAM                       & 20.68          & 43.00             & 56.91          & 16.85         & 42.03          & \textbf{58.68} & 39.69               \\
  KER                       & \textbf{21.87} & \textbf{44.46} & \textbf{58.92} & \textbf{17.37} & \textbf{43.00}          & 58.55          & \textbf{40.70}     \\
  \bottomrule
  \end{tabular}
  }
  \label{mask}
  \end{table}

To validate the effectiveness of our KER module, we conduct comparisons with similar algorithms as detailed in Table \ref{mask}. 
Implicit Relation Reasoning (IRR) \citep{jiang2023cross} utilizes a Masked Language Modeling (MLM) approach, akin to BERT \citep{devlin2018bert}, to implicitly aggregate vision and text features, yielding favorable results. 
However, we argue that the random masking approach employed in IRR may fail to effectively capture key concepts, as if often masks common words like ``is'' and ``a''. 
Predicting such words does not substantially enhance the model's ability to discern nuanced differences in the text, limiting its capacity to focus on more critical, domain-specific information.
By contrast, our KER module explicitly incorporates meaningful keywords, allowing for more precise and fine-grained reasoning.
To further illustrate its superiority, we devise a Masked Attribute Modeling (MAM) module specifically tailored for attribute words, as these terms often provide more discriminative information in retrieval tasks. 
In our experiments, the MAM module and our IRR module achieve 0.46\% and 1.47\% improvements in mR compared with IRR, respectively,
which highlights the importance of key concepts in RSITR task and the effectiveness of KER module.

\subsubsection{Sensitivity analysis of hyperparameter on objective function}
\begin{table}[t!]
  \centering
  \caption{Ablation on hyperparameter on objective function on RSICD Test Set.}
  \resizebox{\linewidth}{!}{
  \begin{tabular}{ccccccccc}
  \toprule
  \multirow{2}{*}{$\gamma$(MLM)} & \multicolumn{3}{c}{Caption Retrieval}    & \multicolumn{3}{c}{Image Retrieval}    & \multirow{2}{*}{mR} \\
                          & R@1            & R@5            & R@10           & R@1            & R@5            & R@10           &                     \\
  \midrule
  0.1                     & 21.23          & 43.55          & 55.17          & 16.51          & \textbf{43.28} & 58.16          & 39.65               \\
  0.2                     & 19.76          & 42.18          & 56.08          & \textbf{17.77} & 43.07          & 58.72          & 39.60               \\
  0.3                     & 20.59          & 43.55          & 58.01          & 17.42          & 42.96          & \textbf{59.01} & 40.26               \\
  0.4                     & 21.77          & 43.28          & 58.54          & 17.26          & 42.87          & 58.23          & 40.33               \\
  0.5                     & \textbf{21.87} & \textbf{44.46} & \textbf{58.92} & 17.37          & 43.00          & 58.55          & \textbf{40.70}       \\
  0.6                     & \textbf{21.87} & 43.37          & 57.37          & 17.09          & 43.09          & 58.72          & 40.25               \\
  0.7                     & 21.42          & 43.54          & 56.81          & 17.11          & 42.58          & 58.76          & 40.04               \\
  0.8                     & 21.59          & 43.73          & 57.18          & 16.71          & 43.04          & 58.81          & 40.18               \\
  0.9                     & 21.13          & 40.71          & 56.27          & 17.53          & 43.06          & 58.99          & 39.62               \\
  1                       & 19.12          & 40.26          & 55.63          & 16.72          & 42.40          & 58.59          & 38.79             \\
  \bottomrule
  \end{tabular}
  }
  \label{MLM}
  \end{table}
  As shown in Table \ref{MLM}, we conduct further investigations into the hyperparameters of the loss function, specifically focusing on the weight of the MLM component while keeping the ratio of global and 
  local loss functions constant. The experimental results indicate that the optimal weight for the MLM loss component is 0.5. 
  It is important to emphasize that the weight assigned to MLM should not be excessively high, as the primary objective of the model remains on the process of contrastive learning, 
  and predicting keywords through MLM serves as a secondary task to support the main alignment process. 
  Overemphasizing the MLM component could detract from the model's ability to capture the nuanced relationships necessary for effective image-text retrieval.

\subsubsection{Impact of the drop epoch and the drop ratio}
\begin{figure}[!t]
  \centering
  \includegraphics[width=\linewidth]{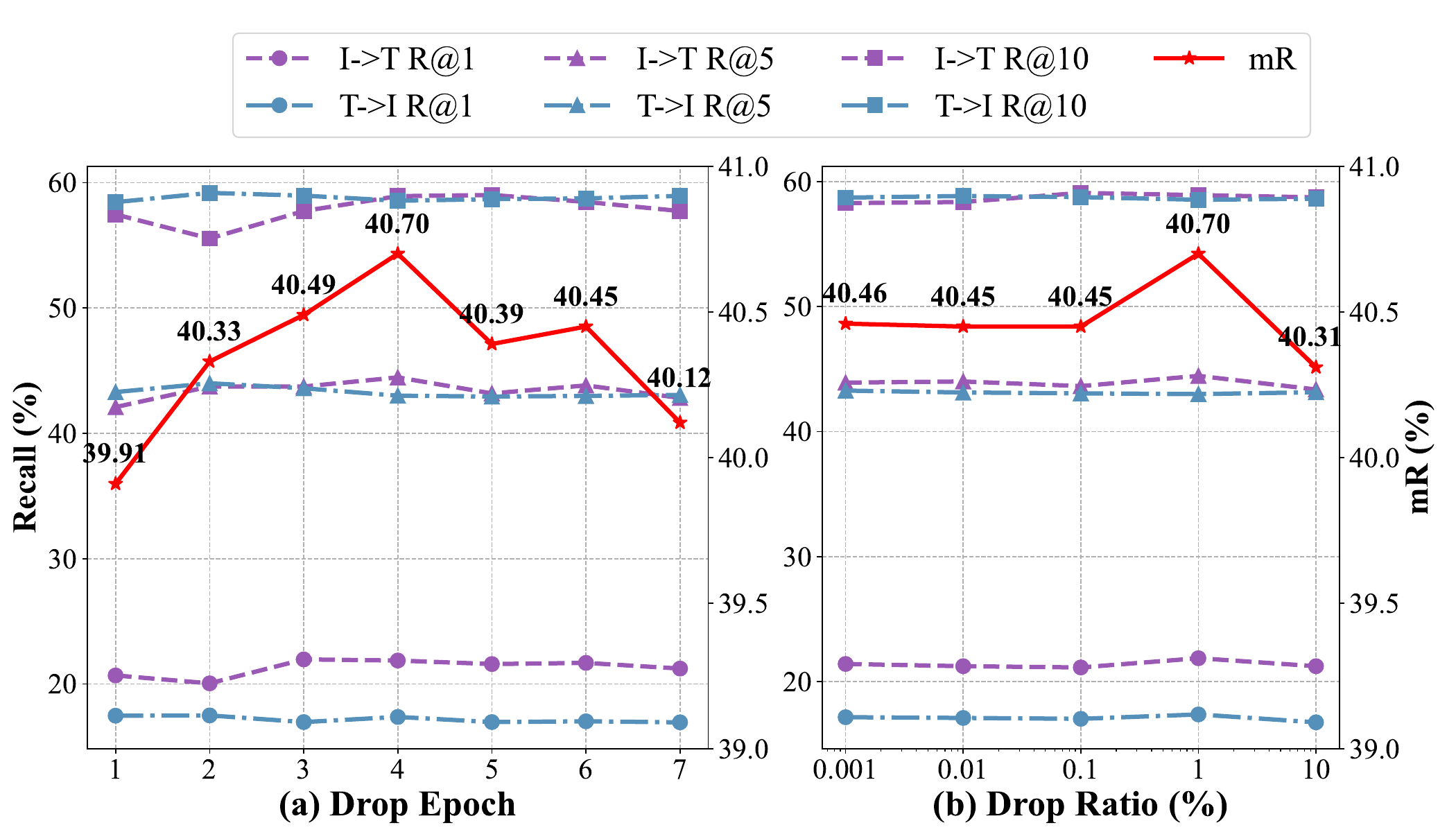}    
  \caption{Impacts of (a) Drop epoch and (b) Drop ratio. Note that R@k (k=1,5,10) refer to the left vertical coordinates while meanR refers to the right vertical coordinates.}
  \label{drop}
\end{figure}
As illustrated in Fig. \ref{drop} (a), we vary the drop epoch $K$ of the EBA strategy from 1 to 7 to explore its impact. Before the designated drop epoch, the model encounters the entire dataset. The results indicate that the optimal drop epoch is the 4th epoch, achieving a 0.79\% improvement in mR. This suggests that dropping data at this stage provides the best balance between data exposure and noise elimination.

Additionally, we examine the impact of the drop ratio in Fig. \ref{drop} (b) to further investigate its role in the EBA strategy. 
The results reveal that a drop ratio of 1\%, meaning that the lowest 1\% of similarity values in the similarity bank are used as the threshold to filter global and local similarities for the next epoch, 
effectively removes noisy sample pairs. A lower drop ratio results in insufficient noise elimination, as too few samples are filtered out. 
A higher drop ratio removes too many normal samples, leading to a decrease in performance. Based on these findings, we determine that a 1\% drop ratio strikes the optimal balance 
between eliminating noise and retaining enough sample diversity for effective training.

\subsubsection{Impact of the quantity of keywords for each dataset}
\begin{table}[!t]
  \centering
  \caption{Impact of the quantity of keywords for each dataset on RSICD Test Set.}
  \resizebox{\linewidth}{!}{
  \begin{tabular}{ccccccccc}
  \toprule
  \multirow{2}{*}{Total} & \multirow{2}{*}{Word} & \multicolumn{3}{c}{Caption Retrieval}    & \multicolumn{3}{c}{Image Retrieval}    & \multirow{2}{*}{mR} \\
                         &                       & R@1            & R@5            & R@10           & R@1            & R@5            & R@10           &                     \\
  \midrule
  27                     & 16                    & 19.67          & 40.90           & 55.17          & 17.05          & 42.27          & 58.24          & 38.88               \\
  57                     & 32                    & 20.59          & 41.99          & 57.18          & 16.98          & 43.81          & \textbf{59.16} & 39.95               \\
  109                    & 64                    & 20.86          & 43.00             & 56.63          & 17.69          & 42.34          & 57.90           & 39.74               \\
  198                    & 128                   & 20.59          & 42.18          & 57.09          & 16.85          & 42.73          & 58.68          & 39.69               \\
  394                    & 256                   & \textbf{22.32} & 43.64          & 56.63          & \textbf{17.82} & \textbf{43.77} & 58.98          & 40.53               \\
  800                    & 512                   & 21.87          & \textbf{44.46} & \textbf{58.92} & 17.37           & 43.00          & 58.55          & \textbf{40.70}     \\
  \bottomrule
  \end{tabular}
  }
  \label{keywords}
\end{table}
We further evaluate the impact of the number of keywords obtained through word frequency analysis.
As shown in Table \ref{keywords}, ``Word''  refers to the number of words selected based on their frequency for each dataset, 
and ``Total'' represents the cumulative number of unique keywords obtained by merging and deduplicating across all three datasets. 
The results suggest that selecting the optimal number of keywords is crucial, as an overly small or excessively large keyword list may hinder performance. 
Selecting too few keywords may miss critical concepts, while too many can introduce unnecessary noise. After experimenting with different ranges, 
we determine that selecting the top 512 most frequent words for each dataset strikes the best balance, ensuring sufficient coverage of key concepts while minimizing noise.

\setlength{\tabcolsep}{5pt}
\begin{table}[!t]
  \centering
  \caption{Ablation experiments with different combinations of training datasets. Note that only the mR metric is reported.}
  \resizebox{\linewidth}{!}{
  \begin{tabular}{ccccccccccccccccccccc}
  \toprule
  \multirow{2}{*}{Model} &  \multicolumn{3}{c}{Datasets} & \multicolumn{3}{c}{Test sets}\\
              &                RSICD    &    RSITMD  & NWPU         &  RSICD    &    RSITMD  & NWPU                \\
  \midrule
  1           &               \checkmark&            &             & 24.11          & 34.65          & 12.55       \\
  2           &                         &  \checkmark&             & 5.71           & 9.50           & 3.03        \\
  3           &                         &            &  \checkmark & 31.57          & 39.11          & 46.86       \\
  4           &               \checkmark&  \checkmark&             & 32.00          & 43.53          & 17.04       \\
  5           &                         &  \checkmark&  \checkmark & 38.20          & 49.22         & \textbf{47.00}        \\
  6           &               \checkmark&            &  \checkmark & 40.11          & 50.39          & 46.95       \\
  7           &               \checkmark& \checkmark &  \checkmark & \textbf{42.07}          & \textbf{53.54}          & 46.83       \\
  \bottomrule
  \end{tabular}
  }
  \label{ablation-datasets}
\end{table}

\setlength\tabcolsep{5pt} 
\begin{table}[!t]
  \centering
  \caption{Trade-off between the mean rank and inference speed. The ``IT'' represents inference time.}
  \resizebox{\linewidth}{!}{
  \begin{tabular}{lcccccc}
  \toprule 
  \multirow{2}{*}{Approach}  & \multicolumn{2}{c}{RSICD} & \multicolumn{2}{c}{RSITMD} & \multicolumn{2}{c}{NWPU} \\
                                                      & mR          & IT(s)       & mR           & IT(s)       & mR         & IT(s)       \\
  \midrule
  VSE++\citep{faghrivse++}                                            & 10.43       & 8.63        & 24.83        & 5.52        & 13.46      & 22.68       \\
  AMFMN\citep{yuan2021exploring}                                         & 16.42       & 25.56       & 29.72        & 6.39        & 32.22      & 148.41      \\
  GaLR\citep{yuan2022remote}                                          & 18.96       & 22.92       & 31.41        & 6.23        & -          & -           \\
  KAMCL\citep{ji2023knowledge}                                       & 23.26       & 11.86       & 36.19        & 5.63        & 40.75      & 28.53       \\
  RemoteCLIP\citep{liu2023remoteclip}                                & 35.26       & 2.42        & 49.38        & 1.42        & 42.90       & 6.38        \\
  EBAKER\citep{ji2024eliminate}                                      & 40.70       & 5.96        & 53.32        & 2.53        &46.02      & 20.30       \\
  iEBAKER(Ours)                                                     & \textbf{42.30}        & 5.96        & \textbf{54.40}         & 2.53        &\textbf{46.83}      & 20.30       \\
  \bottomrule
  \end{tabular}}
  \label{inference}
  \end{table}

\begin{figure*}[!t]
  \centering
  \includegraphics[width=\linewidth]{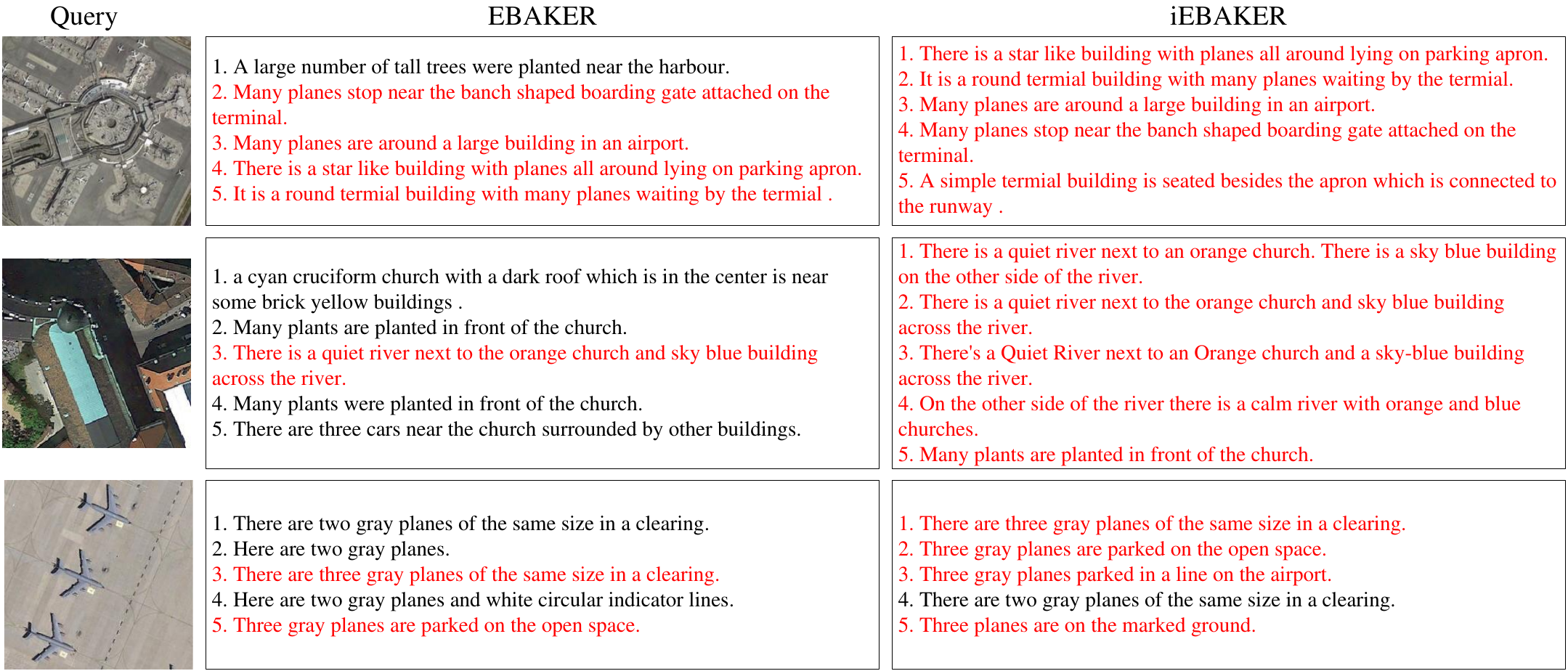}  
  \caption{Visualization of the qualitative caption retrieval results of EBAKER \citep{ji2024eliminate} and our iEBAKER. Each row corresponds to the outcomes obtained from RSICD \citep{lu2017exploring}, RSITMD \citep{yuan2021exploring}, and NWPU \citep{cheng2022nwpu} datasets, respectively. 
  For each image query, the top-5 ranked caption results are displayed, and the matching results are marked as red.}
  \label{visualization-i2t}
\end{figure*}

\begin{figure*}[t]
  \centering
  \includegraphics[width=0.96\linewidth]{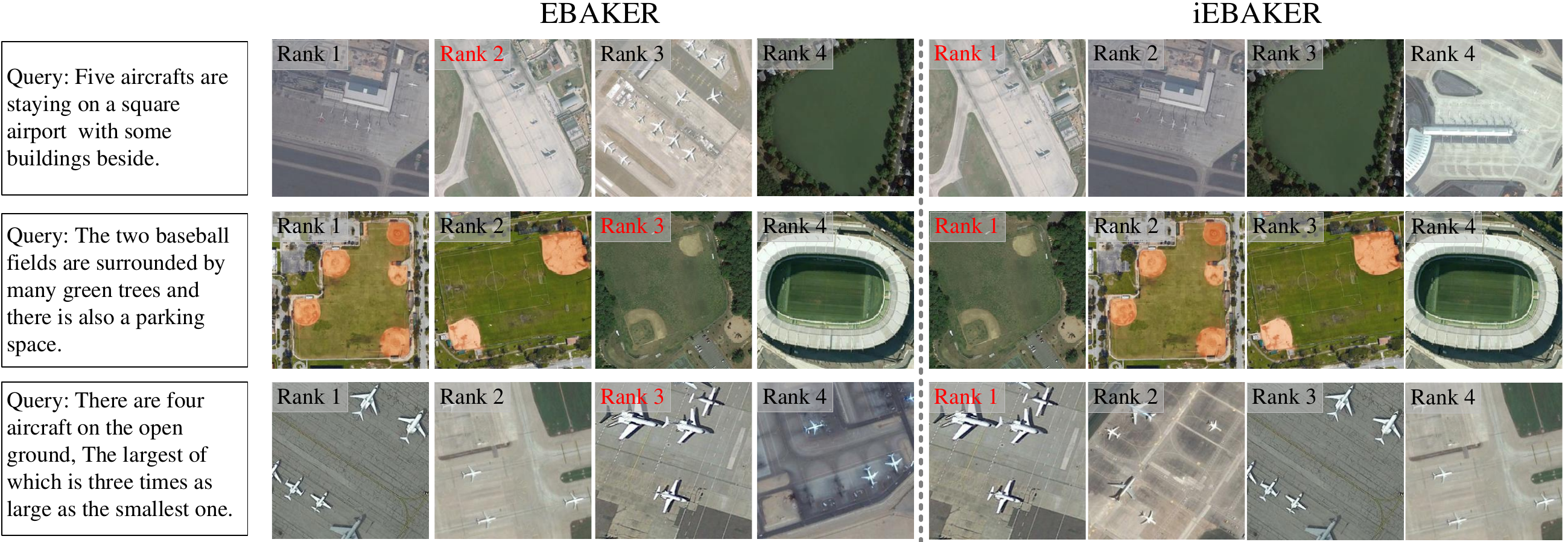}  
  \caption{Visualization of the qualitative image retrieval results of EBAKER \citep{ji2024eliminate} and our iEBAKER. Each row corresponds to the outcomes obtained from RSICD \citep{lu2017exploring}, RSITMD \citep{yuan2021exploring}, and NWPU \citep{cheng2022nwpu} datasets, respectively. 
  For each caption query, the top-4 ranked image results are displayed, and the matching results are marked as red.}
  \label{visualization-t2i}
\end{figure*}

\subsubsection{Impact of different combinations of training datasets}
In this work, we train our model by combining the training sets of the three benchmark datasets followed by \citep{yuan2022remote,zhang2023rs5m}, 
and then test the retrieval performance on different testing sets independently. Thus, we conduct meticulous ablation studies on different combinations of training sets, the  results are shown in Table \ref{ablation-datasets}.
We ensure that all the training parameters are consistent except for the training datasets, and do not perform the SAR processing for fair comparison and simplicity.
Based on the experimental results, we draw four observations and conclusions: (a) more remote sensing training data enables the model possesses better retrieval performance,
this observation is consistent with that of \citep{zhang2023rs5m}; (b) Among the three datasets, the NWPU dataset performs the best, which could be concluded by comparing Models 1 to 3.
(c) A larger training set does not necessarily yield better results, as shown by the NWPU results of Models 5 and 7.
(d) Given that real-world datasets are often incomplete, it is essential to explore methods for transferring knowledge from existing datasets to domains with insufficient data.
Thus, our experiments have paved the way for a new area of research: cross-domain remote sensing image-text retrieval.

\subsection{Trade-off between mean recall and inference speed}
In our evaluation of inference time across various methods, we compare both traditional and FM based approaches on the three benchmark datasets using a single NVIDIA GeForce RTX 4090 GPU. It is important to note that the GaLR \citep{yuan2022remote} method is not replicated on the NWPU dataset due to insufficient details regarding its additional Ppyolo extractor \citep{long2020pp}.
As shown in Table \ref{inference}, the results indicate that our EBAKER and iEBAKER lag behind that of the RemoteCLIP, which relies on excluding global features, in terms of inference speed. This discrepancy arises from the integration of fine-grained local alignment in our model, leading to a requirement for increased inference time. 
Despite this trade-off, our EBAKER and iEBAKER approaches deliver significant improvements in the mR metric, achieving gains of 5.30\%, 3.84\%, and 3.12\% for EBAKER, and 7.04\%, 5.02\%, and 3.93\% for iEBAKER across the datasets, albeit with an increase in inference time by 146\%, 78\%, and 218\%, respectively. When compared with traditional methods such as KAMCL \citep{ji2023knowledge} and GaLR \citep{yuan2022remote}, our approaches exhibit considerable advantages not only in performance but also in inference efficiency. This balance between enhanced retrieval accuracy and manageable inference cost demonstrates the high cost-effectiveness of our method, justifying the additional computational overhead incurred by fine-grained alignment.

\subsection{Visualizations and analyses}
  Figures \ref{visualization-i2t} and \ref{visualization-t2i} display the top-5 qualitative results for caption retrieval and the top-4 qualitative results for image retrieval with our previous version (EBAKER \citep{ji2024eliminate}).

  As depicted, iEBAKER significantly achieves more accurate retrieval results under given queries while EBAKER fails to retrieve them. 
  For the challenge cases that EBAKER fails, iEBAKER successfully retrieves the ground truth images or text captions within the top results. 
  This improvement is largely attributed to the superior image-text embedding space learned by iEBAKER. Specifically, 
  the proposed EBA and SAR strategies enable filter out the weakly correlated sample pairs and mitigate their deviations from optimal embedding space during alignment.
  Compared with EBAKER \citep{ji2024eliminate}, the SAR strategy exhibits a tendency to learn the main visual semantic information with an offline manner.
  For the first example of Fig. \ref{visualization-i2t}, our iEBAKER successfully retrieves the target image in the top ranking position, while EBAKER fails to retrieve it in the first position
  since it incorrectly identifies the ``plane'' as ``tree''. 
  Additionally, the incorrect retrieved results do not mean that they are completely irrelevant to the query. They have the same semantic information as the query.
  For the three examples of Fig. \ref{visualization-t2i}, even the top-2 results of EBAKER are not the correct results, they possess the same semantic information as the query.  
  This suggests that the weakly correlated image-text pairs not only exist in the train set but also in the validation and test sets. 

\section{Conclusion}
In this paper, we have introduced an Improved Eliminate Before Align strategy with Keyword Explicit Reasoning (iEBAKER) framework, designed to facilitate the transfer of FM to RSITRM through a streamlined, one-step fine-grained training. 
We propose an Eliminate Before Align strategy to eliminate weakly correlated pairs, thereby promoting the accuracy of fine-grained contrastive learning.
Besides, this improved version introduce two specific schemes from the perspective of whether local similarity and global similarity affect each other.
We also incorporate a post-processing strategy for optimizing the local and global similarities, and adopts the exponential moving average training scheme for alleviating the
issue of weakly correlated sample pairs. Moreover, we employ a Keyword Explicit Reasoning module, which boosts the discriminative ability by predicting nuanced differences in key concepts.
Finally, the efficacy of our method is rigorously validated through extensive experiments on three widely-used benchmark datasets: RSICD, RSITMD, and NWPU. 

Our method represents a significant advancement by bypassing the RS pretraining stage, offering a viable solution for directly transferring FMs to other tasks within the remote sensing domain.
This approach not only saves a substantial amount of training data typically required across different domains but also opens the door for extending the framework to broader applications, such as product search and pedestrian retrieval.
In the future, we aim to continue exploring methods to to automatically filter data and focus on increasingly fine-grained details, and commit to investigating the 
optimal deployment of multimodal FMs across diverse downstream tasks, ultimately pushing the boundaries of their application potential.

\section*{CRediT authorship contribution statement}
\textbf{Yan Zhang:} Methodology, Software, Formal analysis, Writing - Original Draft, Writing - Review \& Editing. \textbf{Zhong Ji:} Conceptualization, Supervision, Writing - Review \& Editing, Resources, Funding acquisition. \textbf{Changxu Meng:} Conceptualization, Writing - Review \& Editing, Software. \textbf{Yanwei Pang:} Conceptualization, Writing - Review \& Editing. \textbf{Jungong Han:} Methodology, Writing - Review \& Editing.



\section*{Declaration of competing interest}
The authors declare that they have no known competing financial interests or personal relationships that could have appeared to influence the work reported in this paper.

\section*{Acknowledgements}
This work was supported by the National Natural Science Foundation of China (No. 62441235 and No. 62176178).

\bibliographystyle{elsarticle-num-names} 
\bibliography{iEBAKER}




\end{document}